\documentclass{article}

\usepackage[preprint]{corl_2026} 
\usepackage{amsmath, amssymb}
\usepackage{amsthm, graphicx}
\usepackage{booktabs, multirow, xcolor, colortbl}
\usepackage{appendix}
\usepackage{amsmath,amssymb}
\usepackage{graphicx}
\usepackage{xcolor}
\usepackage{dsfont}
\usepackage{mathtools}
\usepackage{mathabx}
\usepackage{amsthm}
\usepackage{wrapfig}
\usepackage{booktabs} 
\usepackage{acronym}
\usepackage{algorithm}
\usepackage{algpseudocode}
\usepackage{todonotes}
\usepackage[font=small]{caption}

\usepackage{multirow}
\usepackage{todonotes}
\title{Online Learning of Robust Legged Odometry with Minimal Exteroceptive Supervision}

\newcommand{\sqrtvins}{\sqrt{\mathrm{VINS}}}
\newcommand{\vVINS}{\boldsymbol{v}^{\mathrm{vins}}}
\newcommand{\vSUP}{\boldsymbol{v}^{\mathrm{sup}}}
\newcommand{\vESN}{\boldsymbol{v}^{\mathrm{esn}}}
\newcommand{\vMeas}{\boldsymbol{v}^{\mathrm{meas}}}
\newcommand{\SigVINS}{\boldsymbol{\Sigma}^{\mathrm{vins}}}
\newcommand{\SigSUP}{\boldsymbol{\Sigma}^{\mathrm{sup}}}
\newcommand{\SigESN}{\boldsymbol{\Sigma}^{\mathrm{esn}}}
\newcommand{\SigMeas}{\boldsymbol{\Sigma}^{\mathrm{meas}}}
\newcommand{\vect}[3]{\prescript{\mathrm{#3}}{}{\boldsymbol{#1}}_{\mathrm{#2}}}

\newcommand{\rotM}[2]{\prescript{\mathrm{#2}}{\mathrm{#1}}{\boldsymbol{R}}}

\acrodef{ESN}{Echo State Network}
\acrodef{VINS}{Visual-Inertial Navigation System}
\acrodef{IMU}{Inertial Measurement Unit}
\acrodef{MLP}{Multi-Layer Perceptron}
\acrodef{RNN}{Recurrent Neural Network}
\acrodef{SGD}{Stochastic Gradient Descent}
\acrodef{IEKF}[InEKF]{Invariant Extended Kalman Filter}
\author{
  Abhijeet M. Kulkarni\\
  University of Delaware\\
  Newark, DE, USA \\
  \texttt{amkulk@udel.edu} \\
  \And
  Yuze Du \\
  University of Delaware\\
  Newark, DE, USA \\
  \texttt{yuzedu@udel.edu} \\
  \And
  Guoquan Huang \\
  University of Delaware\\
  Newark, DE, USA \\
  \texttt{ghuang@udel.edu} \\
}

\begin{document}
\maketitle

\begin{abstract}
    Robust locomotion and navigation for legged robots relies heavily on dependable odometry. Traditional multi-sensor fusion for such state estimation requires meticulous sensor calibration and platform-specific kinematic modeling, which complicates deployment. Industrially packaged exteroceptive sensors can provide accurate motion tracking but remain vulnerable to perceptually degraded conditions. We thus develop a plug-and-play, robust legged odometry system that eliminates the need for explicit exteroceptive-to-proprioceptive calibration or system kinematic modeling. Our approach leverages established exteroceptive motion pipelines as a continuous supervisory signal to train an online learned velocity neural network  directly from proprioceptive data. 
    An Invariant EKF (InEKF) is then used to fuse the learned proprioceptive or exteroceptive velocity (if any) and  IMU data. When exteroception fails due to environmental degradation, the system seamlessly falls back to using the learned proprioceptive model, yielding a resilient legged odometry that readily adapts to new hardware. We demonstrate the platform-agnostic, easily deployable nature of our approach on different quadruped platforms, showcasing promising results in maintaining robust motion estimation across challenging scenarios.
\end{abstract}
\section{Introduction}

Legged robots are becoming increasingly prevalent in real-world applications~\cite{WuEAAI2024} and are equipped with stable low-level controllers~\cite{DiCarlo2018IROS,Kumar2021RSS} that typically utilize
high-rate proprioceptive sensing, such as IMUs and joint sensors, to achieve dynamic locomotion. These platforms are now readily available for purchase and can typically be equipped with high-quality, modular exteroceptive sensors. For these robots to be effective in challenging real-world  applications, they require
robust navigation capabilities for the successful execution of mission-level tasks. One of the fundamental components of such a navigation system is the odometry module, which estimates the robot's ego motion (e.g., position and velocity). Accurate odometry can typically be achieved by leveraging various exteroceptive
sensors, such as cameras~\cite{Peng2025TRO} and LiDARs~\cite{Xu2021RAL}, which often come with a built-in IMU. These industrially packaged sensors are either factory-calibrated or can be calibrated with existing methods~\cite{Furgale2013IROS, Rehder2016ICRA}. However, their performance  can degrade under adversarial conditions, such as low-light conditions for cameras or featureless environments for LiDARs. In contrast, proprioceptive sensors provide information about the robot's base motion through its local kinematics and its interaction with the environment via foot--ground contacts. Yet the effectiveness of proprioceptive sensing is limited by the accuracy of the underlying kinematic and contact models, and fusing proprioceptive with exteroceptive measurements requires accurate inter-sensor modeling and calibration~\cite{Camurri2020Frontiers,Burgul2024IROS}---parameters that can change as the robot's configuration or sensor mounting varies.

In this paper, we propose in-deployment online learning of the proprioception agnostic to hardware platforms, supervised minimally by exteroceptive motion whenever it is available and falling back to the online-learned proprioceptive model when exteroception becomes unreliable. 
The proposed method does not require any pretraining and can therefore be used on any legged robot with proprioceptive sensors, without hand-designed kinematic and contact models. 
In this way, the proposed approach enables a more resilient odometry system that can compensate for adversarial conditions affecting either sensing modality.
In particular, 
the main contributions of this work are the following:
\begin{itemize}
    
    \item We for the first time model the legged proprioception as a \ac{MLP} readout over an \ac{ESN}---a fixed random feature expansion of the proprioceptive history---and train it online with a low-rank EKF  that accounts for both the noise in the proprioceptive measurements and the uncertainty of the exteroceptive supervision. 
    Note that this proprioception is learned  in deployment from sparse exteroceptive supervision, 
    requiring neither pretraining nor hand-designed kinematic and contact models,
    and thus readily transferring to any legged robot only with proprioceptive sensing.
    
    \item We develop a robust \ac{IEKF}-based legged odometry framework that optimally fuses the IMU data and the learned proprioceptive velocity or exteroceptive velocity (if available) through an adaptive measurement selection strategy. 
    
    \item We  validate the proposed method on two quadruped platforms---Boston Dynamics Spot and Ghost Robotics Vision~60---under perception-degraded conditions, demonstrating its resilience compared against proprioceptive and exteroceptive baselines.
\end{itemize}

\section{Related Work}

Perception-based estimators fuse exteroceptive measurements with high-rate inertial and kinematic sensing. Early modular systems such as Pronto combined inertial, kinematic, and exteroceptive cues~\cite{Camurri2020Frontiers}, while tightly coupled methods integrate perception and proprioception more directly:
VILENS fuses visual, inertial, LiDAR, and leg odometry in a factor graph for robustness to degenerate modalities~\cite{Wisth2023TRO}, and EKF-based visual--inertial--leg fusion yields low-latency estimates for dynamic gaits~\cite{Dhedin2023ICRA,Burgul2024IROS}. Purely proprioceptive and contact-aided estimators instead rely on IMU, joint-encoder, and foot-contact measurements~\cite{Hartley2020IJRR,Lin2022CoRL}, but depend on hand-designed
kinematic and contact models that degrade under slip, impacts, terrain compliance, or payload changes. Learning-based methods address this by learning measurement or dynamics models from data: learned inertial odometry produces displacement estimates from IMU histories~\cite{Buchanan2022CoRL}, deep leg-inertial odometry transfers learned proprioceptive models from simulation to real quadrupeds~\cite{Wasserman2025CoRL}, and recent observer and dynamics models for motion planning~\cite{Kulkarni2025arXiv}. 
These models, however, are trained offline and require representative data, motivating our in-deployment, online-learned alternative.

Online learning~\cite{LesortLLMD2020} is attractive for robots operating in dynamic and uncertain environments. Early approaches learned input--output mappings with locally weighted non-parametric models~\cite{Klanke2008JMLR}, later extended to dynamics models for control~\cite{NguyenTuong2008NeurIPS};
both, however, assume access to the full system state. Other methods learn a residual model~\cite{Jiahao2022CoRL} or split learning into offline and online phases~\cite{Okawara2025RAS}. A parallel line of work uses reservoir computing~\cite{Tanaka2019NN}, which extracts features from the input history with a randomly initialized reservoir and trains only a readout layer on top; this has been applied to dynamics learning for trajectory prediction~\cite{Kim2012RAS} and control~\cite{Polydoros2016IROS,Folgheraiter2023IEEEAccess}, again assuming full state access. The standard linear readout is less expressive than a nonlinear one such as an \ac{MLP}~\cite{Jaeger2009CSR}, but nonlinear readouts are costly to train online owing to their many weights.

Kalman filtering has long been used to train neural networks~\cite{Singhal1988NeurIPS}, offering faster convergence than backpropagation but historically limited to small networks because the covariance update scales cubically with the number of parameters. Decoupled EKF variants reduce this cost~\cite{Puskorius1991IJCNN} but discard the correlations between parameters. More recently, low-rank EKF methods~\cite{Chang2023CoLLA} train large networks online while preserving these correlations through a
low-rank approximation. We leverage this method to train the readout of our proprioceptive odometry model online, accounting for both the noise in the proprioceptive measurements and the uncertainty of the exteroceptive supervision.
\section{Efficient ESN-based Proprioception via  Low-Rank EKF}
\label{sec:esn-lofi}

At the core of the proposed legged odometry (see Fig.~\ref{fig:system}) is the \ac{ESN}-based proprioceptive module that is learned online during deployment with sparse exteroceptive motion supervision. 
Specifically, the proprioceptive module produces the fallback velocity measurement $\vESN_t$ with associated uncertainty $\SigESN_t$ from noisy joint-level sensing. 
The history of joint-level signals carries sufficient information to recover the base velocity, but representing this history directly results in an input whose dimension grows with the window length.
A common approach is to use a recurrent feature extractor trained offline on collected data~\cite{Buchanan2022CoRL, Kulkarni2025arXiv},
which is not possible for online plug-and-play deployments under consideration.
We thus turn to reservoir computing and employ an \ac{ESN}~\cite{Sun2024TAI}, which compresses the history
into a reservoir state through fixed, randomly initialized recurrent dynamics and  requires no training of the recurrent weights. 
Note that \ac{ESN}s have been shown to act same as delay-embeddings of histories under certain conditions \cite{Hart2020NN}.


Specifically, at each sampling time $t$, the joint positions $\boldsymbol{q}_t$, velocities $\dot{\boldsymbol{q}}_t$,
and torques $\boldsymbol{\tau}_t$ form the observation $\boldsymbol{o}_t$, and the reservoir state $\boldsymbol{x}_t$ evolves as
\begin{equation}\label{eq:esn}
  \boldsymbol{o}_t =
  \begin{bmatrix}
    \boldsymbol{q}_t^\top & \dot{\boldsymbol{q}}_t^\top & \boldsymbol{\tau}_t^\top
  \end{bmatrix}^\top,
  \qquad
  \boldsymbol{x}_{t+\Delta t} = \mathrm{ESN}(\boldsymbol{x}_t,\, \boldsymbol{o}_t),
\end{equation}
where $\Delta t$ is the sampling interval. 
%
The joint measurements are noisy and so the reservoir state is itself uncertain. 
As a common practice, we assume the observation $\boldsymbol{o}_t$ is Gaussian with covariance $\boldsymbol{\Sigma}_o = \mathrm{blkdiag}(\boldsymbol{\Sigma}_q, \boldsymbol{\Sigma}_{\dot{q}}, \boldsymbol{\Sigma}_{\tau})$, stacking the noise of the joint positions, velocities, and torques, and the reservoir state as Gaussian with covariance $\boldsymbol{P}^x_t$. 
In order to determine the uncertainty of the new reservoir state, we first compute the Jacobians $\boldsymbol{A}_t$ and $\boldsymbol{B}_t$ with respect to $\boldsymbol{x}$ and $\boldsymbol{o}$, evaluated at $(\boldsymbol{x}_t, \boldsymbol{o}_t)$, via automatic differentiation available with PyTorch \cite{Paszke2019NeurIPS},
and then perform covariance propagation as follows:
\begin{equation}\label{eq:esn_cov}
  \boldsymbol{P}^x_{t+\Delta t}
  = \boldsymbol{A}_t \boldsymbol{P}^x_t \boldsymbol{A}_t^\top
  + \boldsymbol{B}_t \boldsymbol{\Sigma}_o \boldsymbol{B}_t^\top.
\end{equation}
Note the echo-state property~\cite{Jaeger2009CSR} keeps the recurrence contractive
and thus  $\boldsymbol{P}^x_t$ remains bounded rather than accumulating without limit.

\subsection{\ac{MLP} Readout Weights}

Because the recurrent weights of the \ac{ESN} are fixed, the reservior state $\boldsymbol{x}_t$ carries both task-relevant and task-irrelevant features. Therefore, for these features to be useful for body velocity prediction, an output mapping is learned online that extracts the components informative of base velocity. We use an MLP $h_{\hat{\boldsymbol{\theta}}_t}$ with online-learned weights $\hat{\boldsymbol{\theta}}_t$ to read out the base velocity from the reservoir state:
\begin{equation}\label{eq:output_mapping}
  \vESN_t = h_{\hat{\boldsymbol{\theta}}_t}(\boldsymbol{x}_t),
\end{equation}
which is the learned proprioceptive velocity measurement and will be used in the \ac{IEKF} to be fused with IMU data (see Section~\ref{sec:leg-odom}).

We now seek to  \emph{online} update the readout weights $\hat{\boldsymbol{\theta}}_t$  in real time using supervision from the extereoceptive (VINS) module whenever available and reliable.
While online learning for neural networks~\cite{LesortLLMD2020} is typically implemented with stochastic-gradient optimizers (i.e., \ac{SGD}, Adam, or variants thereof)~\cite{Paszke2019NeurIPS},
our  online setting make this choice inadequate:
\begin{itemize}
    
\item  \ac{SGD} produces only a point estimate of $\hat{\boldsymbol{\theta}}_t$, with no quantification of uncertainty. Additional steps~\cite{WangTPAMI2026} are therefore needed to make it usable in our pipeline, where the \ac{IEKF}, as the primary estimator, requires a noise characterization for the measurement [see~\eqref{eq:state_meas}].

\item   \ac{SGD}-based approaches treat each sample uniformly and ignore its quality. If the exteroceptive supervisory signal is of poor quality, the readout should be updated cautiously; if it is reliable, more aggressively. Such sample-dependent weighting is in principle possible with \ac{SGD}, but only when the relative
quality of samples can be assessed against an offline-collected dataset~\cite{Song2022TNNLS}, which is unavailable in our online streaming setting.

\item In online learning, with streaming data, the samples are seen one at a time, and the data distribution may shift over time. In this setting, the neural network must by learn extract maximal information from each sample. To achieve this, \ac{SGD} based approaches require to maintain a replay buffer/dataset of past samples~\cite{Rolnick2019NeurIPS, Jiahao2022CoRL} to extract more information from each with backpropagation \cite{Chang2023CoLLA}. This is not only computationally expensive, but also requires additional memory and careful management of the buffer to avoid overfitting to recent samples or forgetting old ones. 

\end{itemize}

These limitations argue for a filtering view~\cite{Singhal1988NeurIPS} of online readout training, in which weights are estimated jointly with their uncertainty, each VINS sample is incorporated by weighing it against the current posterior, and weight evolution between updates is modeled explicitly. 

\subsection{Online Readout Training with Low-Rank EKF}
\label{sec:low-rank}

Following the discussion above, 
we advocate a filtering-based online training for the proposed proprioceptive ESN, by treating the optimal readout parameters
$\boldsymbol{\theta}_t^{\star}$ as a latent state to be estimated sequentially. 
Two signals drive the estimate: 
(i) the reservoir state $\boldsymbol{x}_t$---a fixed encoding of recent proprioceptive history---is the input, modeled as uncertain with covariance $\boldsymbol{P}^x_t$; 
and (ii) the extereoceptive supervisor velocity $\vSUP_t$ with uncertainty as covariance $\SigSUP_t$. 
Online learning then amounts to maintaining an estimate $\hat{\boldsymbol{\theta}}_t$ of $\boldsymbol{\theta}_t^{\star}$---the readout that maps the reservoir state to the true base velocity---together with a quantification of its uncertainty.

In the online setting, we have no prior knowledge of the robot's operating conditions, and hence no parametric model for how
$\boldsymbol{\theta}_t^{\star}$ evolves over time.
We therefore model its evolution as a random walk (which we found empirically reasonable, see Fig.~\ref{fig:weight-cov}). 
This makes the readout adapt to non-stationary conditions---changing terrain, payload, or contact dynamics---while remaining identifiable from extereoceptive supervisory data. 
Together with the supervised velocity measurement, we have the following state-space model:
\begin{equation}\label{eq:ssm}
\begin{aligned}
  \boldsymbol{\theta}_{t+\Delta t}^{\star}
    &= \boldsymbol{\theta}_t^{\star} + \boldsymbol{w}_t,
    & \boldsymbol{w}_t &\sim \mathcal{N}(\boldsymbol{0}, q\boldsymbol{I}), \\
  \vSUP_t
    &= h_{\boldsymbol{\theta}_t^{\star}}(\boldsymbol{x}_t + \boldsymbol{\epsilon}^x_t)
       + \boldsymbol{\epsilon}^v_t,
    & \boldsymbol{\epsilon}^x_t &\sim \mathcal{N}(\boldsymbol{0}, \boldsymbol{P}^x_t),\quad
      \boldsymbol{\epsilon}^v_t \sim \mathcal{N}(\boldsymbol{0}, \SigSUP_t),
\end{aligned}
\end{equation}
where the scalar $q$ sets the adaptation rate: a small $q$ enforces a near-stationary readout, while a larger $q$ allows faster tracking of distribution shifts.

EKF maintains a mean $\hat{\boldsymbol{\theta}}_t$ and covariance $\boldsymbol{P}_t$ on the readout MLP parameters. 
It is important to note that the dimension of these parameters $D = \dim(\boldsymbol{\theta})$ is
large, and thus a full-covariance EKF---storing $\boldsymbol{P}_t \in \mathbb{R}^{D \times D}$ and solving an $O(D^3)$ system
at each update---is impractical for real-time onboard use. 
As such, we adopt the low-rank EKF (LOFI)~\cite{Chang2023CoLLA}, which represents the posterior \emph{precision} in diagonal-plus-low-rank form:
\begin{equation}\label{eq:dlr}
  \boldsymbol{P}_t^{-1} = \boldsymbol{\Upsilon}_t + \boldsymbol{W}_t \boldsymbol{W}_t^{\top},
  \quad \boldsymbol{\Upsilon}_t = \mathrm{diag}(\Upsilon_{t,1}, \ldots, \Upsilon_{t,D}) \in \mathbb{R}^{D \times D},\quad
  \boldsymbol{W}_t \in \mathbb{R}^{D \times L},\quad L \ll D,
\end{equation}
As a result, the posterior is summarized by $(\hat{\boldsymbol{\theta}}_t,
\boldsymbol{\Upsilon}_t, \boldsymbol{W}_t)$ without a dense covariance.
In EKF, we linearize the nonlinear measurement \eqref{eq:ssm} readout about the current estimate through the Jacobians: 
\begin{equation}\label{eq:jacobians}
  \boldsymbol{H}_t = \left.\frac{\partial h_{\boldsymbol{\theta}}(\boldsymbol{x}_t)}{\partial \boldsymbol{\theta}}\right|_{\hat{\boldsymbol{\theta}}_t^{-}},
  \qquad
  \boldsymbol{J}_t = \left.\frac{\partial h_{\hat{\boldsymbol{\theta}}_t^{-}}(\boldsymbol{x})}{\partial \boldsymbol{x}}\right|_{\boldsymbol{x}_t},
\end{equation}
which are obtained by automatic-differentiation, and combine the reservoir-state and extereoceptive uncertainties into a single effective measurement covariance: $\boldsymbol{R}_t = \SigSUP_t + \boldsymbol{J}_t \boldsymbol{P}^x_t \boldsymbol{J}_t^{\top}$. 

The filter proceeds in two steps that operate directly on the factors of \eqref{eq:dlr}. 
Propagation under the zero-mean random walk leaves the mean unchanged and inflates the covariance by $q\boldsymbol{I}$, applied to the precision factors. The update incorporates the innovation
$\boldsymbol{r}_t = \vSUP_t -
h_{\hat{\boldsymbol{\theta}}_t^{-}}(\boldsymbol{x}_t)$ by appending the measurement directions $\boldsymbol{H}_t^{\top}\boldsymbol{R}_t^{-1/2}$ to $\boldsymbol{W}_t$ and restoring rank $L$ via an SVD truncation, with the mean correction obtained through the Woodbury identity, see \cite{Chang2023CoLLA} for details. This keeps the gain and
posterior update at $O\!\big(D(L+C)^2\big)$ time and $O(DL)$ memory, where $C = 3$ is the velocity dimension---\emph{linear} in the number of readout weights. 
The updated readout produces the proprioceptive velocity
$\vESN_t = h_{\hat{\boldsymbol{\theta}}_t}(\boldsymbol{x}_t)$, with covariance $\SigESN_t = \boldsymbol{H}_t \boldsymbol{P}_t \boldsymbol{H}_t^{\top} + \boldsymbol{J}_t \boldsymbol{P}^x_t \boldsymbol{J}_t^{\top}$, formed cheaply from the diagonal-plus-low-rank precision factors via the Woodbury identity. The resulting module (Alg.~\ref{alg:readout_update})
learns online while accounting for uncertainty on both sides of the update---the proprioceptive input $(\boldsymbol{x}_t, \boldsymbol{P}^x_t)$ and the exteroceptive supervisor $(\vSUP_t, \SigSUP_t)$---and supplies the velocity estimate $(\vESN_t, \SigESN_t)$ whenever required to the primary estimator \ac{IEKF} described next.
\begin{algorithm}[t]
\caption{Online Update of Readout Weights with Supervision}
\label{alg:readout_update}
\begin{algorithmic}[1]
\Require $(\hat{\boldsymbol{\theta}}_{t-\Delta t}, \boldsymbol{\Upsilon}_{t-\Delta t}, \boldsymbol{W}_{t-\Delta t})$,
  $(\boldsymbol{x}_t, \boldsymbol{P}^x_t)$, $(\vSUP_t, \SigSUP_t)$, $q$
\Ensure $(\hat{\boldsymbol{\theta}}_t, \boldsymbol{\Upsilon}_t, \boldsymbol{W}_t)$,
  $(\vMeas_t, \SigMeas_t)$
\Statex
\State $(\hat{\boldsymbol{\theta}}_t^{-}, \boldsymbol{\Upsilon}_t^{-}, \boldsymbol{W}_t^{-}) \gets
  \textsc{Predict}(\hat{\boldsymbol{\theta}}_{t-\Delta t}, \boldsymbol{\Upsilon}_{t-\Delta t}, \boldsymbol{W}_{t-\Delta t}, q)$
\State $(\vESN_t, \boldsymbol{\Sigma}^{\mathrm{esn}}_t, \boldsymbol{R}_t, \boldsymbol{H}_t, \boldsymbol{J}_t) \gets
  \textsc{PredictOutput}(\hat{\boldsymbol{\theta}}_t^{-}, \boldsymbol{\Upsilon}_t^{-}, \boldsymbol{W}_t^{-}, \boldsymbol{x}_t, \boldsymbol{P}^x_t, \SigSUP_t)$
\State $s_t \gets \textsc{SwitchingManager}(\SigESN_t, \SigSUP_t)$
\If{$s_t = \textsc{unusable}$} \Comment{Supervision unreliable then readout drives \ac{IEKF}}
  \State $(\hat{\boldsymbol{\theta}}_t, \boldsymbol{\Upsilon}_t, \boldsymbol{W}_t) \gets
    (\hat{\boldsymbol{\theta}}_t^{-}, \boldsymbol{\Upsilon}_t^{-}, \boldsymbol{W}_t^{-})$
  \State $(\vMeas_t, \SigMeas_t) \gets (\vESN_t, \boldsymbol{\Sigma}^{\mathrm{esn}}_t)$
\Else \Comment{Supervision reliable then it drives \ac{IEKF} and updates the readout}
  \State $(\hat{\boldsymbol{\theta}}_t, \boldsymbol{\Upsilon}_t, \boldsymbol{W}_t) \gets
    \textsc{LofiUpdate}(\hat{\boldsymbol{\theta}}_t^{-}, \boldsymbol{\Upsilon}_t^{-}, \boldsymbol{W}_t^{-}, \boldsymbol{H}_t, \vSUP_t, \vESN_t, \boldsymbol{R}_t)$
  \State $(\vMeas_t, \SigMeas_t) \gets (\vSUP_t, \SigSUP_t)$
\EndIf
\State \textbf{return} $(\hat{\boldsymbol{\theta}}_t, \boldsymbol{\Upsilon}_t, \boldsymbol{W}_t)$,
  $(\vMeas_t, \SigMeas_t)$
\end{algorithmic}
\end{algorithm}
%

\section{Robust \ac{IEKF}-based Legged Odometry}
\label{sec:leg-odom}
\begin{figure}[h]
  \centering
  \includegraphics[width=.85\linewidth]{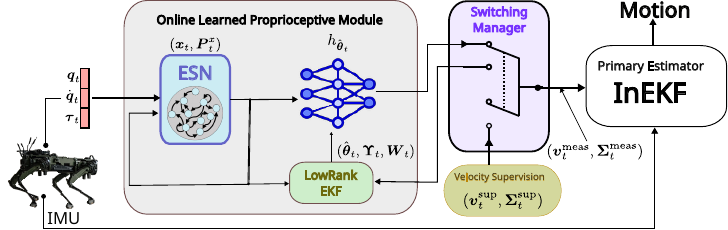}
  \caption{The proposed legged odometry system. 
  The \ac{IEKF} propagates with IMU and updates with velocity measurements  supplied by either the VINS supervisor or the learned proprioceptive module, selected by the switching manager. 
  While VINS is reliable, if needed, it additionally supervises online learning of the proprioception.}
  \label{fig:system}
\end{figure}
We now optimally fuse the learned proprioceptive velocity (and exteroceptive, if any) and IMU data within the \ac{IEKF} \cite{Hartley2020IJRR, Lee2025IROS}
to provide  high-rate, robust motion estimates to enable resilient locomotion and navigation.
As shown in Fig.~\ref{fig:system},
the proposed \ac{IEKF}   propagates the navigation state with the IMU and updates with a velocity measurement supplied by one of two interchangeable sources—an exteroceptive supervisory VINS  or a learned proprioception—selected at runtime by a switching manager. 
Note that when VINS is reliable, if needed, it may additionally act as a supervisor, providing the target signal for online learning of the proprioceptive module.

Specifically, the based navigation state $\boldsymbol{\chi}_t \in \mathrm{SE}_2(3)$  packs the orientation ${\rotM{I}{G}}_t$, velocity ${\vect{v}{I}{G}}_t$, and position ${\vect{p}{I}{G}}_t$ of the base frame $\{I\}$ in the global frame $\{G\}$, 
as well as the IMU accelerometer and gyroscope biases $\boldsymbol{b}_t \in \mathbb{R}^6$.
These states  propagate with the strapdown IMU  and are corrected by the local velocity measurements $\vMeas_t \in \mathbb{R}^3$, which admits a {\em right-invariant} output:
\begin{equation}
  \boldsymbol{X}_t = (\boldsymbol{\chi}_t,\, \boldsymbol{b}_t), \quad
  \boldsymbol{\chi}_t = \left[
  \begin{smallmatrix}
    {\rotM{I}{G}}_t & {\vect{v}{I}{G}}_t & {\vect{p}{I}{G}}_t \\
    \boldsymbol{0}_{1\times3} & 1 & 0 \\
    \boldsymbol{0}_{1\times3} & 0 & 1
  \end{smallmatrix}\right], \quad
  \boldsymbol{b}_t = \left[
  \begin{smallmatrix} \boldsymbol{b}_{a,t} \\ \boldsymbol{b}_{g,t} \end{smallmatrix}\right], \quad
  \vMeas_t = \bigl({\rotM{I}{G}}_t\bigr)^{\!\top} {\vect{v}{I}{G}}_t + \boldsymbol{\nu}_t,
  \label{eq:state_meas}
\end{equation}
where $\boldsymbol{\nu}_t$ is the zero-mean measurement noise $\mathcal{N}(\boldsymbol{0}, \SigMeas_t)$.



Besides the learned proprioceptive velocity as explained in Section~\ref{sec:esn-lofi}, we here briefly discuss the off-the-shelf extereoceptive VINS velocity. 
%
By tightly coupling visual and inertial data, the VINS module estimates base motion, from which we obtain body-frame velocity measurements $\vVINS_t$ and the associated estimate uncertainty $\SigVINS_t$. It plays two roles: (i) as a measurement source for the \ac{IEKF} when exteroception is reliable, and (ii) as the uncertainty-aware supervisory signal $(\vSUP_t, \SigSUP_t)$ for online learning of the proprioceptive module. 
In particular, we adopt the recent $\sqrtvins$~\cite{Peng2025TRO}, whose square-root filter formulation improves numerical stability, reduces the memory and computation burden, and preserves a positive semi-definite covariance---a property required by both the primary estimator and the online training of the proprioceptive module.
It is important to note that VINS can become unreliable when its visual front end degrades---for example, under textureless scenes, or rapid lighting changes/ no light---in which case the proposed legged odometry system falls back to the learned proprioceptive module for the velocity measurement. 




To ensure robust update, we introduce a switching manager that governs 
the VINS and proprioceptive velocity data paths: 
(i) $\vMeas_t \leftarrow \vVINS_t$ when the VINS estimate is reliable and the ESN covariance is large, 
and (ii) $\vMeas_t \leftarrow \vESN_t$ when the VINS estimate is unreliable and the ESN covariance dropped to a low-level during online-learning as in \cite{Joshi2023ICRA}. 
%
Specifically, the reliability of the VINS module is monitored by computing an overall system health score that combines feature-tracking quality and velocity estimate covariance into a single normalized value using configurable weights. The resulting health score is classified into healthy or unusable, based on preset thresholds. Once VINS velocity estimates marked unusable and ESN covariance is small, VINS outputs will no longer be IEKF measurement source nor supervisory signal for proprioceptive module. See Algorithm~\ref{alg:readout_update} for switching manager governed proprioceptive-ESN readout update and details on the switching manager can be found in Appendix~\ref{sec:switching_appendix}.



\section{Experimental Validation}

We evaluate the proposed legged odometry approach on two quadrupedal platforms: (i) the Boston Dynamics Spot and (ii) the Ghost Robotics Vision~60, focusing on odometry performance under exteroceptive-supervision outages. For Spot, we use a publicly available dataset with ground-truth pose~\cite{Noh2025arXiv} and manually toggle the velocity supervision off in controlled windows, so that the online-learning component can be evaluated in isolation from the measurement-selection mechanism. The supervisory velocity is derived from the available ground-truth pose, which also serves as the reference for evaluating the estimated trajectories.

The Vision~60 evaluation, by contrast, is conducted in a realistic deployment. An external Intel RealSense D435i is mounted on the robot as the exteroceptive sensor, with its embedded IMU driving the \ac{IEKF} and no extrinsic calibration performed between the RealSense and the robot's proprioceptive sensors. $\sqrtvins$ provide velocity supervision using the attached Realsense camera. The robot traverses visually degraded zones, and the complete framework Figure~\ref{fig:system} must work to produce robust odometry for the camera's IMU frame. Since ground-truth pose is unavailable, we assess the result by overlaying the estimated trajectory on the floor plan of the building.

We implement Algorithm~\ref{alg:readout_update} in PyTorch~\cite{Paszke2019NeurIPS}. Readout updates run at the supervision rate: 10\,Hz on Spot (the dataset reference rate) and 15\,Hz on Vision~60 (the camera rate). The \ac{ESN} reservoir has dimension 1024, with weights drawn from LeCun's normal distribution and scaled to a spectral radius of 0.99. The readout \ac{MLP} has architecture $1024 \to 8 \to 3$ with $\tanh$ activation, resulting in $D = 8227$ parameters including biases. The LO-FI precision approximation uses rank $L = 64$. On an NVIDIA RTX~2000~Ada GPU, the ESN uncertainty propagation \eqref{eq:esn_cov} takes 0.43\,ms on average and each readout weight update takes 6.3\,ms---both well within the proprioceptive and exteroceptive supervision rates, confirming that the method runs in real time.

The Spot evaluation targets the online-learning  of the proprioceptive module (Section~\ref{sec:esn-lofi}). We first numerically validate the random-walk assumption on the readout MLP parameters~\eqref{eq:ssm}. To this end, we train the MLP readout with full supervision on the complete \textsc{Atrium} sequence. As shown in Figure~\ref{fig:weight-cov}, after an initial learning phase the weights stabilize and remain nearly constant---
consistent with stationary optimal weights $\boldsymbol{\theta}^\star_t$ and providing empirical support for the random-walk model.
\subsection{Evaluations on Spot Dataset}
\begin{wrapfigure}{r}{0.65\linewidth}
    \centering
    \includegraphics{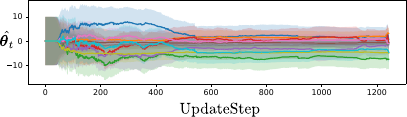}
    \caption{Evolution of 10 randomly chosen (out of $D = 8227$) readout MLP
    weights trained with full ground-truth velocity supervision on the
    \textsc{Atrium} sequence from the Spot dataset~\cite{Noh2025arXiv}.}
    \label{fig:weight-cov}
    \vspace{-5pt}
\end{wrapfigure}

The readout MLP converges quickly, as evident from Figure~\ref{fig:weight-cov}, suggesting that partial supervision suffices to learn a reliable map from the \ac{ESN} state to the robot's base velocity. We therefore evaluate the effect of limiting supervision to a fixed window and then relying on the learned module as the velocity source for the \ac{IEKF} (Section~\ref{sec:leg-odom}). Specifically, we test supervision windows of 10\,s, 30\,s, and 60\,s, and compare against two baselines: GaRLILEO~\cite{Noh2025arXiv}, an exteroceptive--proprioceptive odometry system, and a contact-aided Leg-\ac{IEKF}~\cite{Hartley2020IJRR} with offline-calibrated kinematics and heuristic-based contact detection. Table~\ref{tab:spot_dataset_results} reports the detailed comparison across baselines and supervision levels.

Our method's accuracy improves with longer supervision windows: within 30\,s of supervision it learns a usable map for sequences whose subsequent unsupervised motion stays close to what was seen during training. One of the exceptions is \textsc{Overpass}, where the supervision window covers only flat-ground walking and stair ascent but the unsupervised portion includes a stair descent that lies outside the training distribution, resulting in larger drift. On most of the remaining sequences, where the unsupervised motion remains within the distribution of the supervision window, our method performs competitively with the exteroceptive--proprioceptive GaRLILEO and even surpasses the Leg-\ac{IEKF}, despite using no offline calibration or pretraining---highlighting the
plug-and-play nature of the approach.

\begin{table*}[t]
\centering
\caption{Results on Spot Dataset at different supervision levels and comparisons with baselines.}
\label{tab:spot_dataset_results}
\small
\setlength{\tabcolsep}{2pt}

\begin{minipage}[t]{0.49\textwidth}
\centering
\resizebox{\linewidth}{!}{%
\begin{tabular}{l|lccccc|}
\toprule
Sequence & Method & APE$_{\text{trans}}$ & RPE$_{\text{trans}}$ & APE$_{\text{rot}}$ & RPE$_{\text{rot}}$ & Sup. \% \\
 & & [m] & [m] & [deg] & [deg/m] & [\%] \\
\midrule
\textsc{Atrium} & GaRLILEO & 0.82 & 0.06 & 1.72 & 0.55 & -- \\
109.93 m & Leg-\ac{IEKF} & 2.27 & 0.11 & 6.46 & 0.91 & -- \\
124.50 s & Ours-10s & 3.90 & 0.13 & 3.23 & 0.53 & 8.00 \\
 & Ours-30s & 1.68 & 0.10 & 3.22 & 0.54 & 24.10 \\
 & Ours-60s & 0.90 & 0.06 & 3.17 & 0.54 & 48.20 \\
\midrule
\textsc{BiCorridor} & GaRLILEO & 1.43 & 0.06 & 5.52 & 0.89 & -- \\
240.82 m & Leg-\ac{IEKF} & 3.25 & 0.14 & 9.13 & 1.41 & -- \\
277.29 s & Ours-10s & 4.81 & 0.15 & 6.83 & 0.77 & 3.60 \\
 & Ours-30s & 3.66 & 0.13 & 6.81 & 0.83 & 10.80 \\
 & Ours-60s & 3.37 & 0.14 & 6.88 & 0.85 & 21.70 \\
\midrule
\textsc{BridgeLoop} & GaRLILEO & 1.19 & 0.08 & 2.72 & 1.06 & -- \\
161.17 m & Leg-\ac{IEKF} & 1.24 & 0.18 & 6.48 & 3.44 & -- \\
187.20 s & Ours-10s & 3.46 & 0.20 & 7.62 & 0.95 & 5.30 \\
 & Ours-30s & 2.52 & 0.17 & 7.59 & 0.97 & 16.00 \\
 & Ours-60s & 1.96 & 0.16 & 7.62 & 1.02 & 32.10 \\
\midrule
\textsc{CorriLoop} & GaRLILEO & 1.63 & 0.07 & 5.68 & 0.74 & -- \\
208.68 m & Leg-\ac{IEKF} & 3.30 & 0.12 & 10.84 & 1.13 & -- \\
229.40 s & Ours-10s & 4.78 & 0.12 & 5.36 & 0.77 & 4.40 \\
 & Ours-30s & 1.31 & 0.09 & 5.30 & 0.75 & 13.10 \\
 & Ours-60s & 1.73 & 0.09 & 5.27 & 0.75 & 26.20 \\
\midrule
\textsc{Downstair} & GaRLILEO & 3.92 & 0.10 & 3.42 & 1.08 & -- \\
233.75 m & Leg-\ac{IEKF} & 8.66 & 0.14 & 9.92 & 1.26 & -- \\
270.90 s & Ours-10s & 19.18 & 0.47 & 6.42 & 0.68 & 3.70 \\
 & Ours-30s & 32.21 & 0.68 & 6.04 & 0.65 & 11.10 \\
 & Ours-60s & 37.14 & 0.66 & 6.58 & 0.63 & 22.20 \\
\bottomrule
\end{tabular}%
}
\end{minipage}
\hfill
\begin{minipage}[t]{0.49\textwidth}
\centering
\resizebox{\linewidth}{!}{%
\begin{tabular}{l|lccccc}
\toprule
Sequence & Method & APE$_{\text{trans}}$ & RPE$_{\text{trans}}$ & APE$_{\text{rot}}$ & RPE$_{\text{rot}}$ & Sup. \% \\
 & & [m] & [m] & [deg] & [deg/m] & [\%] \\
\midrule
\textsc{Overpass} & GaRLILEO & 1.53 & 0.09 & 4.04 & 1.23 & -- \\
169.17 m & Leg-\ac{IEKF} & 43.27 & 1.52 & 27.23 & 5.87 & -- \\
213.49 s & Ours-10s & 264.58 & 1.54 & 17.18 & 0.74 & 4.70 \\
 & Ours-30s & 61.23 & 0.60 & 20.61 & 0.69 & 14.10 \\
 & Ours-60s & 45.00 & 0.54 & 4.90 & 0.67 & 28.10 \\
\midrule
\textsc{Quad} & GaRLILEO & 7.35 & 0.08 & 3.36 & 0.84 & -- \\
447.83 m & Leg-\ac{IEKF} & 27.62 & 0.11 & 22.74 & 0.90 & -- \\
503.69 s & Ours-10s & 53.60 & 0.55 & 10.04 & 0.50 & 2.00 \\
 & Ours-30s & 30.73 & 0.48 & 10.73 & 0.60 & 6.00 \\
 & Ours-60s & 11.48 & 0.11 & 9.39 & 0.55 & 11.90 \\
\midrule
\textsc{SlopeStair} & GaRLILEO & 2.36 & 0.06 & 2.81 & 1.05 & -- \\
273.37 m & Leg-\ac{IEKF} & 13.38 & 0.12 & 20.63 & 1.20 & -- \\
307.49 s & Ours-10s & 6.48 & 0.14 & 8.67 & 0.79 & 3.30 \\
 & Ours-30s & 5.87 & 0.13 & 8.47 & 0.86 & 9.80 \\
 & Ours-60s & 6.08 & 0.12 & 8.80 & 0.83 & 19.50 \\
\midrule
\textsc{Tunnel} & GaRLILEO & 3.52 & 0.08 & 2.85 & 0.44 & -- \\
247.94 m & Leg-\ac{IEKF} & 8.56 & 0.10 & 9.76 & 0.82 & -- \\
277.00 s & Ours-10s & 4.42 & 0.10 & 7.09 & 0.40 & 3.60 \\
 & Ours-30s & 6.10 & 0.12 & 8.46 & 0.43 & 10.80 \\
 & Ours-60s & 4.98 & 0.12 & 6.84 & 0.44 & 21.70 \\
\midrule
\textsc{Upstair} & GaRLILEO & 1.50 & 0.07 & 4.05 & 0.93 & -- \\
197.22 m & Leg-\ac{IEKF} & 2.01 & 0.17 & 8.07 & 2.89 & -- \\
227.89 s & Ours-10s & 7.00 & 0.20 & 7.15 & 0.96 & 4.40 \\
 & Ours-30s & 3.52 & 0.18 & 7.07 & 0.94 & 13.20 \\
 & Ours-60s & 2.06 & 0.16 & 7.13 & 0.93 & 26.30 \\
\bottomrule
\end{tabular}%
}
\end{minipage}
\end{table*}


\subsection{Evaluations on Vision60}
\begin{figure}
    \centering
    \includegraphics{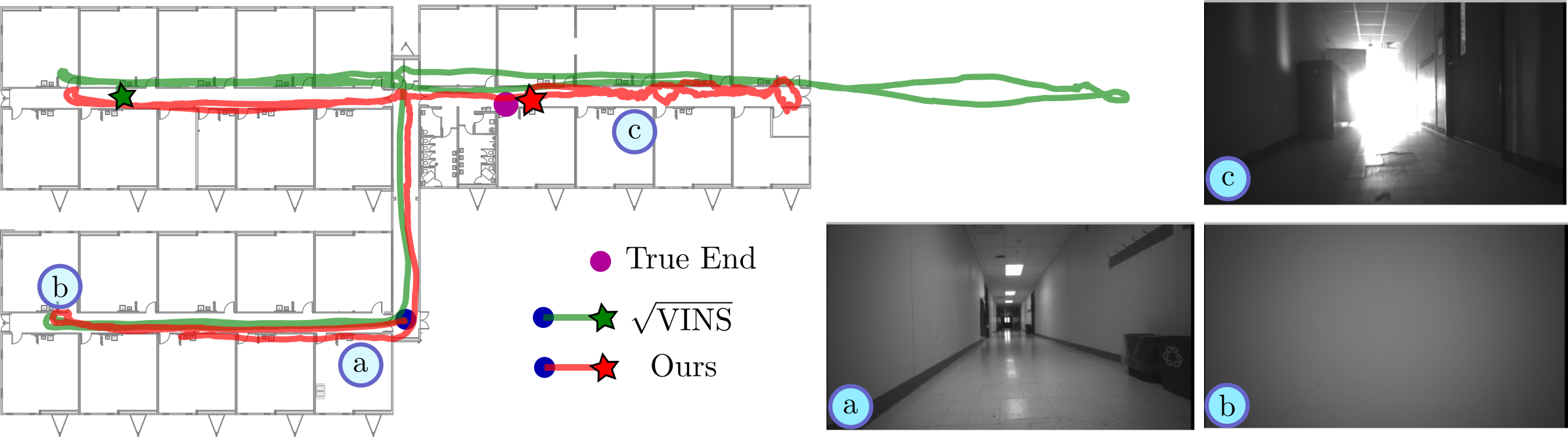}
    \caption{Hallway exploration with lighting degradation. }
    \label{fig:starlab_walking}
    \vspace{-8pt}
\end{figure}
\begin{wrapfigure}{r}{0.45\textwidth}
    \centering
    \includegraphics[width=\linewidth]{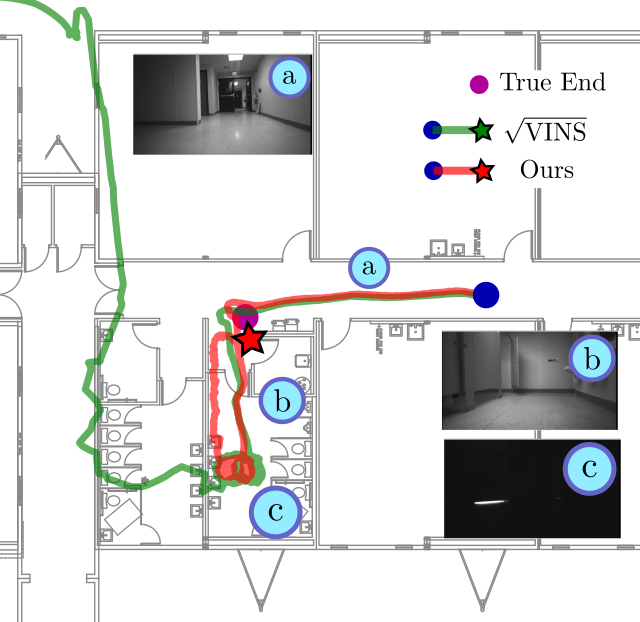}
    \caption{Lavatory exploration with the lights turned off mid-spin.}
    \label{fig:lavatory_exploration}
    \vspace{-12pt}
\end{wrapfigure}

We next test the full pipeline (Figure~\ref{fig:system}) in scenarios where $\sqrtvins$ based velocitiy supervision fails. Figures~\ref{fig:starlab_walking} and~\ref{fig:lavatory_exploration} compare the trajectories from VINS (green)
and our system (red) during hallway and lavatory exploration under conditions
representative of VINS failure modes. The endpoints of the two trajectories
are marked with a green and a red star respectively, and the true endpoint
with a purple dot.

In Figure~\ref{fig:starlab_walking}, the robot starts at waypoint $a$ in a
well-lit hallway, giving Algorithm~\ref{alg:readout_update} ample time to
train the readout MLP. At waypoint $b$, the robot reaches the end of the
hallway, faces a textureless wall, and turns around; VINS briefly loses track
of visual features but recovers. At waypoint $c$, the hallway lights are off
and a bright window at the far end creates strong backlighting, severely
degrading the visual conditions. VINS struggles to extract reliable features,
inflating its covariance and causing odometry drift; the switching manager
detects this and switches the \ac{IEKF} to the learned proprioceptive module
as the velocity source. As a result, the VINS trajectory overshoots the true
endpoint by a considerable margin, while our robust \ac{IEKF} pipeline ends much closer to it.

In Figure~\ref{fig:lavatory_exploration}, the robot again starts in the
well-lit hallway (waypoint $a$). At waypoint $b$, it enters the lavatory, and
after reaching waypoint $c$ it spins in place while the lights are turned off
mid-spin. With no visible features to track, VINS drifts severely; our
approach remains accurate thanks to the preemptive switch to the learned
proprioceptive measurements. As in the hallway sequence, the endpoint
estimated by our approach is much closer to the true end than VINS's drifted
estimate.

\section{Conclusions}
In this work, we proposed an efficient \ac{ESN}-based proprioceptive module whose \ac{MLP} readout is trained on the fly with a low-rank EKF, while systematically accounting for uncertainty in both the proprioceptive sensing and the velocity supervision. Building on this module, we developed a robust \ac{IEKF}-based legged odometry framework that learns the proprioception-to-base-velocity mapping online from extereoceptive derivedvelocity supervision, requiring no prior training or calibration. We evaluated the framework on two quadrupedal platforms and demonstrated our odometry remains robust to visual degradation by falling back to the online-learned proprioceptive module. The approach performs well under sparse online supervision and produces accurate odometry competitive with proprioceptive and perception-based baselines.

\section{Limitations}
Our framework rests on the assumption that exteroception and proprioception do not degrade simultaneously. The learned readout MLP itself is not expected to generalize beyond the distribution covered by its velocity supervision, predicting base velocity for different from supervision motion will be incorrect.

\bibliography{libraries/extra,libraries/rpng,libraries/legged,libraries/related_perception,libraries/res_comp}
\begin{appendices}
\appendixpage
\appendices
\section{Switching Manager}
\label{sec:switching_appendix}

The system health monitor computes a normalized health score \(H \in [0,1]\) as a weighted combination of MSCKF feature quality, SLAM feature quality, and covariance quality:
\begin{equation}
H =
w_{\mathrm{msckf}} s_{\mathrm{msckf}}
+ w_{\mathrm{slam}} s_{\mathrm{slam}}
+ w_{\mathrm{cov}} s_{\mathrm{cov}},
\end{equation}
where \(w_{\mathrm{msckf}} = 0.45\), \(w_{\mathrm{slam}} = 0.20\), and \(w_{\mathrm{cov}} = 0.35\). The MSCKF and SLAM sub-scores are computed from their feature counts using a clipped linear ramp:
\begin{equation}
s_{\mathrm{feat}} =
\operatorname{clip}
\left(
\frac{N - N_{\mathrm{ok}}}
{N_{\mathrm{good}} - N_{\mathrm{ok}}},
0, 1
\right),
\end{equation}
with \((N_{\mathrm{ok}}, N_{\mathrm{good}}) = (5,20)\) for MSCKF features and \((0,10)\) for SLAM features. The covariance sub-score penalizes large velocity uncertainty by first computing
\begin{equation}
\sigma_{\max}
=
\max\left(
\sigma_{v_x},
\sigma_{v_y},
\sigma_{v_z}
\right),
\end{equation}
and then evaluating
\begin{equation}
s_{\mathrm{cov}} =
\operatorname{clip}
\left(
1 -
\frac{
\sigma_{\max} - \sigma_{\mathrm{good}}
}{
\sigma_{\mathrm{bad}} - \sigma_{\mathrm{good}}
},
0, 1
\right),
\end{equation}
where \(\sigma_{\mathrm{good}} = 0.05~\mathrm{m/s}\) and \(\sigma_{\mathrm{bad}} = 0.30~\mathrm{m/s}\). Thus, low velocity uncertainty receives a high covariance score, while uncertainty above the bad threshold contributes zero. The raw score may be smoothed using a moving average over \(10\) samples, and the resulting health state $s_t$ is classified as healthy when \(H \geq 0.70\), degraded when \(0.40 \leq H < 0.70\), and unusable when \(H < 0.40\).

\end{appendices}\end{document}